\pdfoutput=1

\documentclass[letterpaper, 10 pt, conference]{ieeeconf}  

\IEEEoverridecommandlockouts                              
\usepackage{graphics} 
\usepackage[namelimits]{amsmath} 
\usepackage{amssymb}
\usepackage{amsfonts} 
\usepackage{graphicx}
\usepackage{booktabs}
\usepackage{bm}
\usepackage{multirow}
\usepackage{makecell}
\usepackage{hyperref}
\usepackage{threeparttable}
\hypersetup{hidelinks,
	colorlinks=true,
	allcolors=black,
	pdfstartview=Fit,
	breaklinks=true}

\title{\LARGE \bf
Yaw-Guided Imitation Learning \\for Autonomous Driving in Urban Environments
}

\author{
Yandong Liu$^{\dag}$,
	Chengzhong Xu$^{\ddag}$,
	Hui Kong$^{*}$
	\thanks{$\dag$ Center for Cloud Computing, Shenzhen Institutes of Advanced Technology, Chinese Academy of Sciences, Shenzhen, China. Shenzhen College of Advanced Technology, University of Chinese Academy of Sciences, Shenzhen, China, (e-mail: yd.liu@siat.ac.cn).
	}
	\thanks{$\ddag$ The State Key Laboratory of Internet of Things for Smart City (SKL-IOTSC), Department of Computer Science, University of Macau, Macau, (e-mail: czxu@um.edu.mo).}
	\thanks{$*$ The State Key Laboratory of Internet of Things for Smart City (SKL-IOTSC), Department of Electromechanical Engineering (EME), University of Macau, Macau, (e-mail: huikong@um.edu.mo).}
}

\begin{document}

\maketitle
\thispagestyle{empty}
\pagestyle{empty}

\begin{abstract}
Existing imitation learning methods suffer from low efficiency and generalization ability when facing the road option problem in an urban environment. In this paper, we propose a yaw-guided imitation learning method to improve the road option performance in an end-to-end autonomous driving paradigm 
in terms of the efficiency of exploiting training samples and adaptability to changing environments. Specifically, the yaw information is provided by the trajectory of the navigation map. Our end-to-end architecture, Yaw-guided Imitation Learning with ResNet34 Attention (YILRatt), integrates the ResNet34 backbone and attention mechanism to obtain an accurate perception. It does not need high-precision maps and realizes fully end-to-end autonomous driving given the yaw information provided by a consumer-level GPS receiver. By analyzing the attention heat maps, we can reveal some causal relationship between decision-making and scene perception, where, in particular, failure cases are caused by erroneous perception. We collect expert experience in the Carla 0.9.11 simulator and improve the benchmark CoRL2017 and NoCrash. Experimental results show that YILRatt has a 26.27\% higher success rate than the SOTA CILRS. The code, dataset, benchmark and experimental results can be found at \hyperlink{https://github.com/Yandong024/Yaw-guided-IL.git}{https://github.com/Yandong024/Yaw-guided-IL.git}
\end{abstract}


\section{INTRODUCTION}

Autonomous driving has gained much interest as an essential application of artificial intelligence, from industry to academia \cite{9046805}. The modular approach, which incorporates perception, localization, planning, and control techniques, is widely used in industry because of its interpretability \cite{zablocki2021explainability}. In the event of a failure, the module where the defect is located may be analyzed and identified. However, due to the complexity of autonomous driving tasks, the development and maintenance costs of any of the technologies in the module are extremely high \cite{Sun_2020_CVPR}. Therefore, an end-to-end imitation learning (IL) approach has recently become a popular research topic in academia \cite{tampuu2020survey}. This method learns expert experience through deep neural networks \cite{ grigorescu2020survey}. The environment perceived by the sensors is input into the neural network, and the neural network outputs control variables.

\begin{figure}[!htpb]
  \centering
  \includegraphics[scale=0.55]{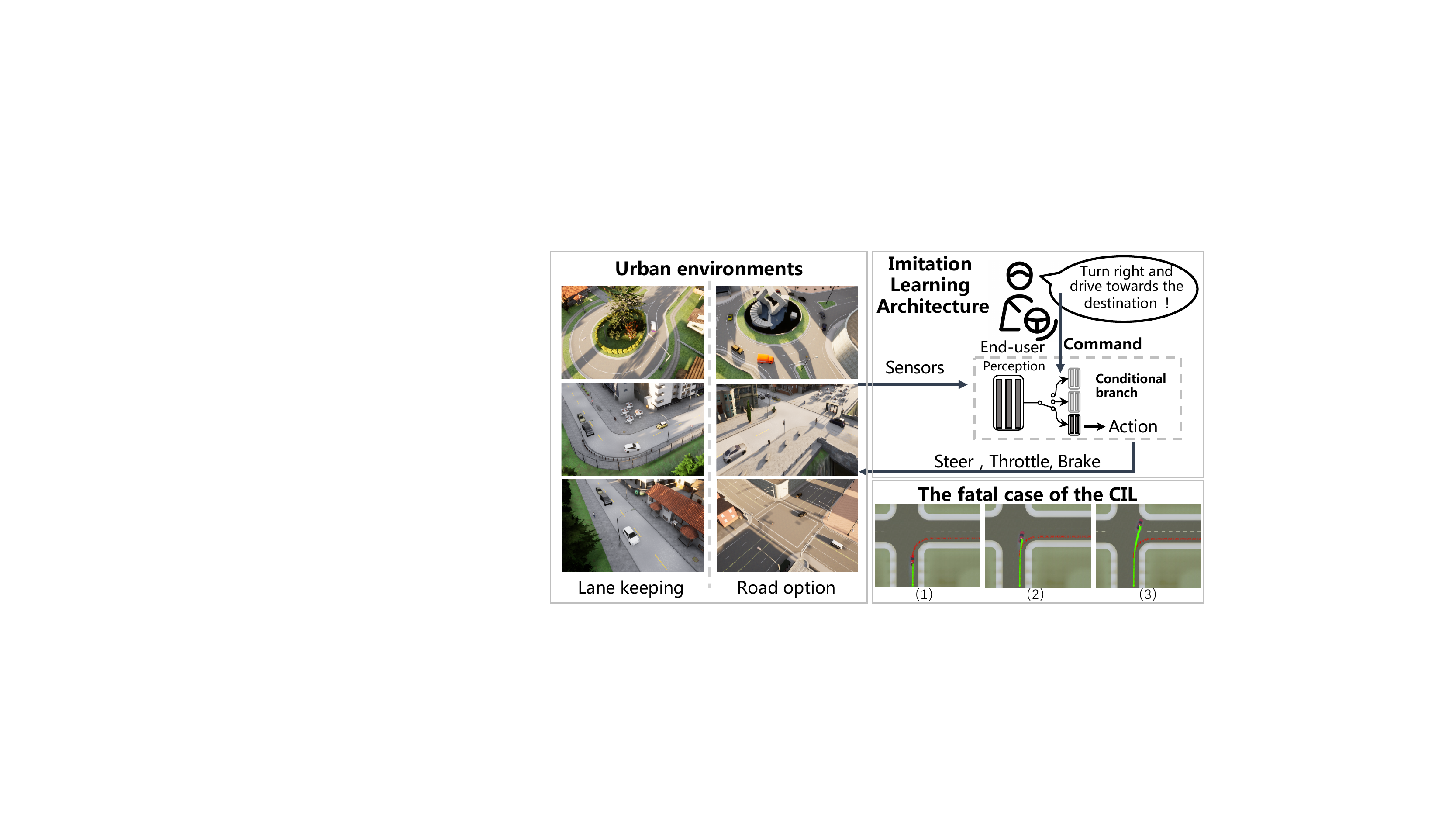}
  \caption{The CIL selects the road by the end-user's command. In training, we only use the turning data of lane following to train the model. In testing, images shown in (1)-(3) demonstrate the intermediate trajectory when taking right turn given by an road option command by end-users. However, the vehicle controlled by CIL cannot make a right turning successfully at the intersection. }
  \label{fig:end-to-end}
\end{figure}

Imitation learning has been successfully applied to lane keeping (following) and obstacle avoidance by learning driving strategies through extensive human driving experience \cite{bojarski2016end}\cite{ muller2006off}. In a lane keeping scene, there is a single mapping relationship between driving behavior and environment. Therefore, the imitation learning strategy can control vehicles to turn left, turn right and go straight. However, in the urban environments, road option become a major obstacle to applying imitation learning. Road option, that is, when a vehicle is at an intersection, T-shape junction or roundabout, the vehicle needs to decide whether to turn left, right or go straight, as shown in Figure \ref{fig:end-to-end}. In such cases, scene and vehicle behavior cannot be modeled by a single mapping. As a result, vehicles cannot select a direction only by sensing the surrounding environment using a camera or LiDAR, and distinguishing the turn of lane keeping from that of road option is a big challenge in applying imitation learning in urban environments.

To deal with this issue, the conditional imitation learning (CIL) method \cite{codevilla2018end} exploits human commands to select the corresponding branch network by executing a commmand of an end-user, where each individual branch network of CIL is trained based on a large amount of image-control pairs sampled accordingly. 
Generally, when the distribution of sampled training data is unbalanced, the branch network parameters cannot be well optimized for the specific road option that lacks enough training samples. As shown in the example in Figure \ref{fig:end-to-end}, we use the turning data of lane keeping to train the model and apply it to the case of the right turning of road option. The vehicle cannot complete the task at the intersection during the test. From this example, we observe that the model trained from the right-turning data of lane following cannot be adapted well to the road option branch. In another word, the CIL does not have enough generalization ability. 

Besides, the complexity of scenes is another major challenge for applying imitation learning in urban environments. Due to weather, traffic signals, and dynamic obstacles, urban environments are extremely complicated \cite{Codevilla_2019_ICCV}. Therefore, a good perception module is needed to provide accurate scene representation and feature extraction for the decision network. Especially, it would be more valuable if the learned features are helpful to reveal the causal relationship between scenes and decisions.

In this paper,  we propose a fully end-to-end imitation learning architecture for autonomous driving, Yaw-guided Imitation Learning with ResNet34 Attention (YILRatt),  where the yaw information is derived from the planned trajectory information instead of the end-user's command.  Due to the yaw guidance, our method has more powerful adaptation ability, and can generalize well to both the turning cases of road option and lane-following scenarios.  In this sense, the method improves efficiency of data utilization (the experimental part Section \ref{section:road option} for details). The network input is an RGB-image captured at each moment. The perception network uses ResNet34 as the backbone, and the attention mechanism strengthens the image feature area, weakens the chaotic area, and enhances the perception ability. Trajectory yaw and vehicle speed are used as measurement inputs to the fully connected network. YILRatt has achieved a $26.27\%$ higher success rate than Conditional Imitation Learning ResNet (CILRS), testing on improved benchmark CoRL2017 and NoCrash, respectively. By analyzing the attention heat maps, we can reveal some causal relationship between decision-making (steer, throttle, and brake) and scene perception, where, in particular, failure cases are caused by erroneous perception. 


\section{Related work}\label{section2:related work}

Standard imitation learning obtains the control strategy through collected expert experience. Usually, the expert data set $\mathcal{D}$ is composed of observation-behavior pairs $\left\langle o_{t}, a_{t}\right\rangle$ at time $t$. Similar to the training process of supervised learning, the observation data are the input of the network $N(o; \theta)$, and the behavior data are the labels. Network parameters are optimized by minimizing the loss function $\left(\ref{equ:loss}\right)$ of prediction and expert behavior. 

\begin{equation}
    \underset{\theta}{\operatorname{minimize}} \sum_{i} \mathcal{L}\left(Network\left(o_{i} ; \theta\right), a_{i}\right)
    \label{equ:loss}
\end{equation}

However, imitation learning  assumes that environmental observations and behaviors in expert data satisfy a single mapping relationship, i.e., $a_{i}=E(o_i)$. When enough data are collected, supervised learning can be used to obtain an approximate function of the expert strategy. For example, imitation learning is successful in tasks such as lane keeping and obstacle avoidance given enough collected data. However, in urban scenes, unmanned vehicles face the task of non-single mappings. The expert behavior at this time is not only determined by environmental observations but also related to the destination's location, i.e., $a_{i}=E(o_i, \boldsymbol{y_i})$, where $\boldsymbol{y_i}$ is a vector related to road option. We choose the yaw information, which is derived from planned trajectory, just in front of the vehicle as the $\boldsymbol{y_i}$. Therefore, the collected data becomes $ \mathcal{D} = \{\left\langle o_{i}, \boldsymbol{y_i}, a_{i}\right\rangle\}^N_{i=1}$. The objective function is adjusted to $\left(\ref{equ:obj}\right)$.

\begin{equation}
    \underset{\theta}{\operatorname{minimize}} \sum_{i} \mathcal{L}\left(Network\left(o_{i}, \boldsymbol{y_i}; \theta\right), a_{i}\right)
    \label{equ:obj}
\end{equation}

The very early work of imitation learning \cite{pomerleau1989alvinn} used a three-layer network to learn human driving behavior to accomplish lane keeping. In recent years, with the development of deep learning, models have stronger environmental perception ability. Imitation learning regained a new life. Especially, \cite{bojarski2016end} applies imitation learning to real-world self-driving scenarios. Primitive imitation learning network architectures can only handle single mapping scenarios. Conditional Imitation Learning (CIL) uses branching networks to solve the problem of non-single mapping at the intersection \cite{codevilla2018end}. The design of branching networks allows human and unmanned vehicles to interact. However, as shown in Figure \ref{fig:end-to-end}, different road option data can only train corresponding model branches, which leads to a poor generalization ability and low data-utilization efficiency. GPS coordinate guidance offers another way of road option \cite{Hecker_2018_ECCV}. The unprocessed GPS coordinate information cannot provide a better representation for the planned route, which increases the difficulty of learning. In response to the shortcomings of the above two methods, we propose trajectory yaw guidance for the road option, where the yaw information derived from the trajectory waypoints is exploited as the input to the network to guide the vehicle to select the correct road.

To improve the end-to-end imitation learning control accuracy, researchers have done a lot of work on both the perception module and the affordance information. Using only RGB images as input for the perception network, CILRS \cite{Codevilla_2019_ICCV} improves the accuracy of perception in complex scenes by using ResNet34's strong feature extraction capability.  \cite{Hecker_2018_ECCV} creates a multi-camera system that can offer data for a 360-degree view of the vehicle's surroundings. Furthermore, \cite{hecker2020learning} uses semantic segmentation technology to improve the perception of the environment. \cite{9197408} uses semantic segmentation technology while adding geometry and motion with computer vision. Multi-sensor fusion technology is an important method to compensate for the shortcomings of a single sensor. \cite{9165167} fuses RGB-image with the depth information provided by lidar. The analysis shows that multi-modal perception is better than single modality. \cite{9119447} also uses RGB-camera and lidar, and network parameters are optimized by a loss function including the semantic segmentation result. On the other hand, the affordance approach obtains the environmental features directly through vehicle-road cooperation and human-computer interaction. Conditional Affordance Learning (CAL) \cite{pmlr-v87-sauer18a} provides more effective high-dimensional information for network by adding affordances such as traffic lights’ status and the vehicle’s distance from the lane centerline. Advice from passengers is used as input information to the network \cite{Kim_2019_CVPR}. The method provides a way for passengers to interact with unmanned vehicles. At the same time, the control strategy in complex scenarios (to avoid pedestrians traffic accidents) is optimized. The bird-view image provides the road and other vehicle information around the unmanned vehicle to optimize the strategy \cite{pmlr-v100-chen20a}.

Imitation learning is based on expert experience data to train a model. Therefore, the amount and distribution of collected data become the strategy bottleneck. Various urban scenes are designed to provide rich data sources for imitation learning \cite{cai2020learning}. Data aggregation technology improves the generalization ability of the model \cite{Prakash_2020_CVPR}. Adding perturbations to the expert experience and penalizing fatal scenarios with a loss function increases the robustness of the model \cite{bansal2018chauffeurnet}. We use the Carla simulator to collect about 200 thousand expert data to train the model. Furthermore, we analyze the distribution of data based on road option and weather. At last, in the Section \ref{section:road option}, experiments have shown that the model based on yaw guidance has strong generalization ability under extreme data distribution conditions.

\section{YILRatt Architecture}\label{section3:YILRatt}

\begin{figure}[!ht]
\centering
\includegraphics[scale=0.48]{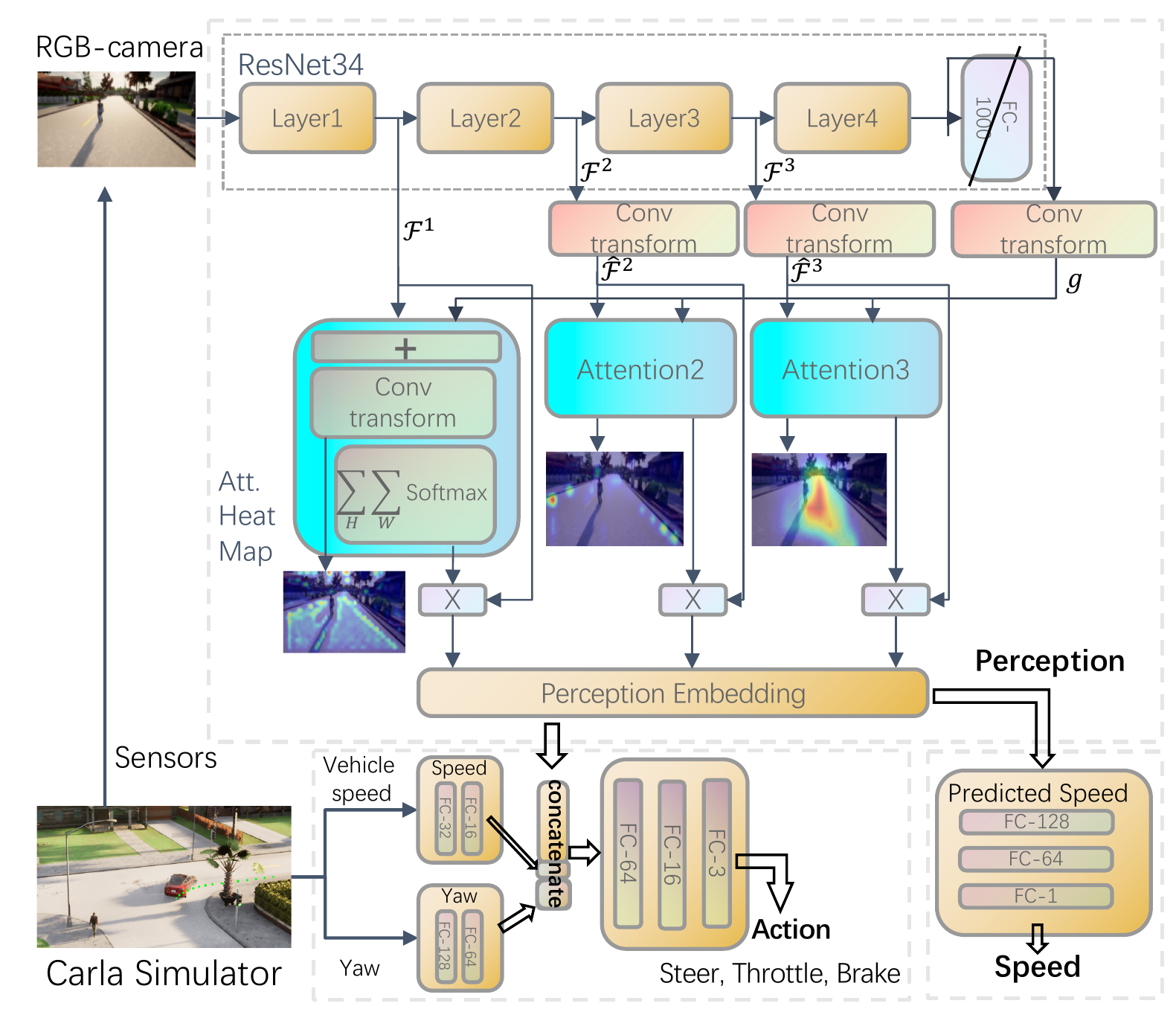}
\caption{Our proposed architecture, called YILRatt, consists of perception, speed, and action modules. The perception module combines ResNet34 with a linear attention mechanism to extract environmental features from a single frame of RGB image. The action module fuses features, measurements and trajectory yaws to predict the action. Speed module improves the accuracy of longitudinal prediction.}
\label{fig:network_architecture}
\end{figure}

The end-to-end imitation learning architecture YILRatt consists of perception $P$, action $A$, and speed $S$ modules, as shown in Figure \ref{fig:network_architecture}.  RGB images $o_i$ taken by the onboard camera taken at each time instance is input into the perception module. The output of the module is the environmental feature representation $P(o_i)$. The vehicle speed $s$ is obtained by the Carla API. The yaw $\boldsymbol{y}$ is calculated by the global trajectory. The action module predicts the behavior of the vehicle by concatenating the $P(o_i)$, $s$ and $\boldsymbol{y}$. The behavior includes $steer \in [-1, 1]$, $throttle \in [0, 1]$, and $brake \in [0, 1]$. The speed module minimizes the error between the predicted speed and the vehicle speed by supervised learning. The loss of the speed module is added to the total loss function (\ref{equ:ary_loss}) to improve the accuracy of longitudinal prediction. We obtain an expert approximate strategy by optimizing the YILRatt’s parameter $ \theta_P, \theta_A, \theta_S$.

\begin{equation}
    \begin{split}
    \alpha * \mathcal{L}( A(s_i, \boldsymbol{y_i}, P(o_i; \theta_P); \theta_A), a_i)+\\ \beta* \mathcal{L}(S(P(o_i; \theta_P);\theta_S), s_i)
    \label{equ:ary_loss}
    \end{split}
\end{equation}

\subsection{Yaw Guidance}

\begin{figure}[!ht]
\centering
\includegraphics[scale=0.8]{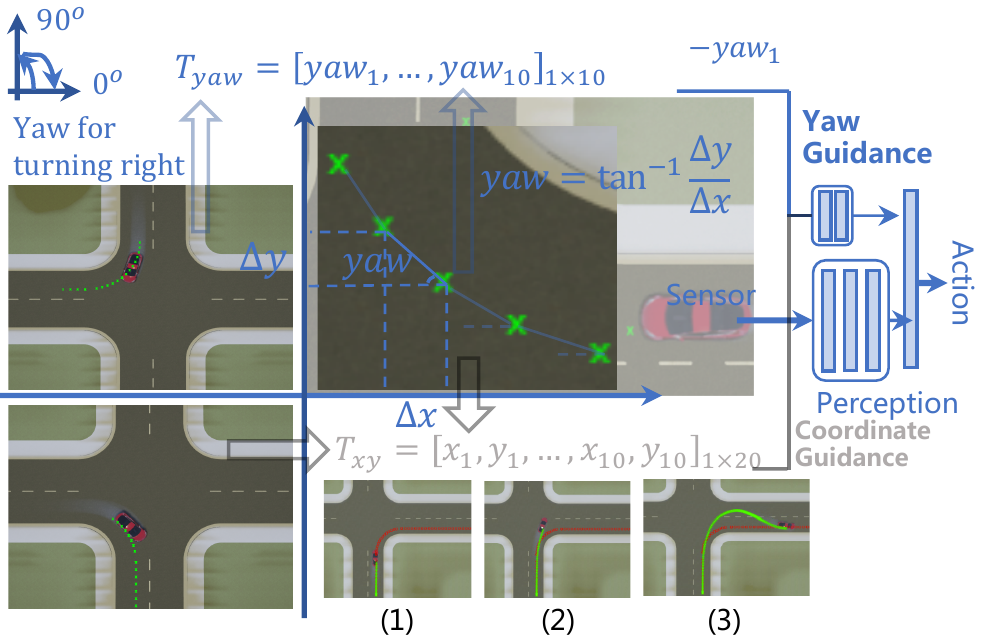}
\caption{Through the global planning of the trajectory, the yaw angle is calculated. The vehicle is guided to select the correct road and drives to the destination.}
\label{fig:yaw_cal}
\end{figure}

An assumption of the method is that the destination is known. Moreover, the global trajectory is planned to ensure that the unmanned vehicle obtains the yaw in the road option. In an urban scenario, it is a reasonable assumption for an end-to-end method for autonomous driving \cite{codevilla2018end}. The guidance information can be either the coordinates or the yaw of the trajectory. 
\cite{Hecker_2018_ECCV}  uses the absolute coordinates of the vehicle as guidance information, which results in a location-specific vehicle control network instead of a road-option (direction choice) one. This means that the network learned by using specific-scene training data can be applied to the same scene in general. Therefore, using absolute GPS coordinates as guidance will result in poor generalization ability in road option. In contrast, the guidance yaw information in our method is a relative quantity, and the yaw guidance can map left (right) turns to the same control amount even in different quadrants (Figure \ref{fig:yaw_cal}), where the left turn (right turn) of different scenarios has the same guidance information. As a result, our method has a more generalization power than the method using absolute quantities as guidance information, e.g. \cite{Hecker_2018_ECCV}. In another words, our method is more efficient in data utilization where we can achieve better end-to-end autonomous driving performance with much fewer training samples. Experiments in Section \ref{section:road option} can show that yaw guidance is more accurate, and it is easier for our road option network to select the road direction successfully. 

By Carla API, we can get yaw from trajectory waypoints. To satisfy the needs of the real world applications, we obtain the yaw information based on the trajectory coordinates. The coordinate difference between two adjacent waypoints on the trajecotry is calculated, and then its arc tangent is calculated and set as the yaw angle in Figure \ref{fig:yaw_cal}. $\boldsymbol{T_{yaw}}$ is a vector composed of yaw information of multiple consecutive waypoints. Thus, the yaw guidance, $\boldsymbol{y}=\boldsymbol{T_{yaw}}- yaw_1$, for road option can be obtained. The guidance method not only achieves the road option, but also improves the model's generalization ability. 
The model trained on the steering data of the lane-following scenarios can be applied to the steering scenario of the road option, and vice versa.
For example, with the same setting as the failure case of conditional imitation learning in Figure \ref{fig:end-to-end}, Figure \ref{fig:yaw_cal} (1)-(3) show that our yaw-guided imitation learning successfully completes the right turn.

\subsection{Perception Module}


We use the ResNet34 as the backbone of the perception module. Using ResNets, the gradient can flow directly from the back layer to the initial filter through the jump connection. The disappearance of the gradient of the deep network is solved, thereby improving the accuracy of the image recognition. However, the visual reasoning of ResNets in environment understanding is largely difficult to understand, hindering the understanding of success and failure. Especially in application scenarios with demanding safety requirements such as autonomous driving, we need to understand the causal relationship between decisions and scenarios. Therefore, we introduce the attention mechanism to explain the causal relationship by attention heat maps.

The core idea of the attention in this paper is to combine the local features of the middle layer and the global features of the output layer to strengthen the salient area and suppress the information chaotic area \cite{xu2015show}. ResNet34 has the local feature $\mathcal{F}^{l}=\left\{f_{1}^{l}, f_{2}^{l}, \ldots, f_{n}^{l}\right\}$ features in the layer $l \in\{1,2, \ldots, L\}$. $f_{n}^{l}$ is the output vector at spatial location $i$ of $n$ total spatial locations. The global feature vector of ResNet34 is $\bm{g}$. $f_{n}^{l}$ and $\bm{g}$ are two arguments of the same dimension of the compatible scoring function (\ref{equ:compatible}) . 

\begin{equation}
C_{i}=\left\langle f_{i}^{l}+g, \theta_{att.}\right\rangle, \quad i \in\{1, \cdots, n\}
\label{equ:compatible}
\end{equation}

We can simplify the two arguments to an addition operation by element-wise. $\theta_{att.}$ can be trained to obtain corresponding features related to the driving scene. The output of the compatible function is the set of scores $C\left(\hat{\mathcal{F}}^{l}, \bm{g}\right)=\left\{c_{1}^{l}, c_{2}^{l}, \cdots_{n}^{l}, c_{n}^{l}\right\}$, where $\hat{\mathcal{F}}^{l}$ is the feature of $\mathcal{F}^{l}$ under a linear mapping of the $f_{n}^{l}$ to the dimensionality of $\bm{g}$. After being normalized by softmax, the score $c_{n}^{l}$  is used as the weight of the local feature. The results of different layers are concatenated to provide environmental features for the action and speed network.

\section{Experiments}\label{section4:experiment}
In this section, we present the experimental setup, results and analysis. Firstly, we compare the Yaw-guided imitation learning with other road option methods in a small-scale static ``square'' scene. The success rate of turning shows that the data utilization of the Yaw-guided method is high. Next, in the urban environment Carla Town01 and Town02, we compare different architectures in benchmark. Then, by analyzing the experimental metrics such as success rate, lane violation and traffic light violation, we show that YILRatt is the SOTA end-to-end imitation learning architecture. Finally, the causes of failure cases are analyzed by attention heat maps. 

\subsection{Yaw-guided Road Option in the Static ``Square'' Scene} \label{section:road option}

\begin{figure}[!ht]
\centering
\includegraphics[scale=0.49]{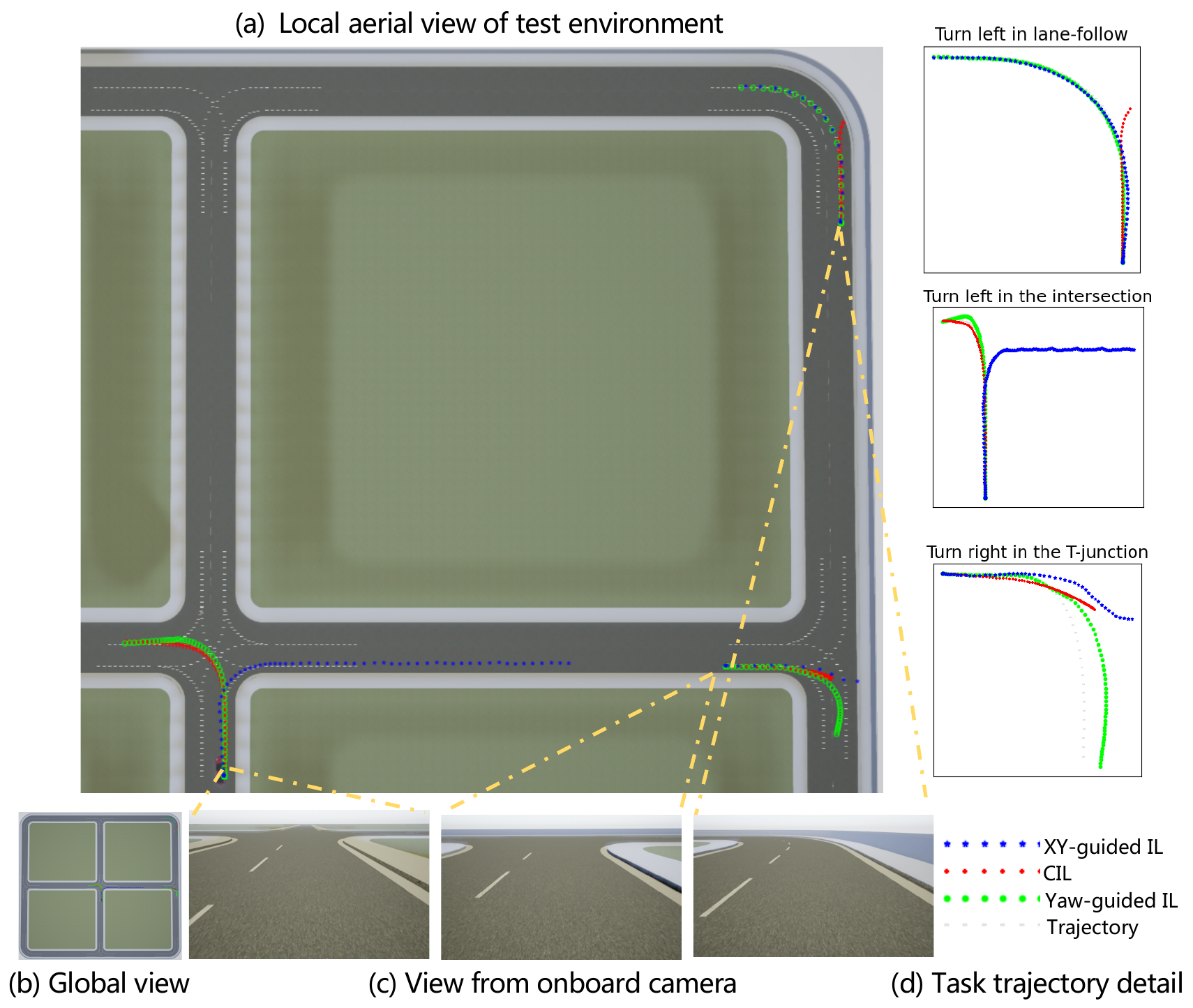}
\caption{Comparison trajectory of three road option methods (XY-guided IL, Yaw-guided IL, CIL) and standard Trajectory. (a) and (b) are local and global bird's-eye views of the test scene. (c) is the view from onboard camera, that is, the single RGB-image inputs into the network. (d) are trajectory details of the three turning tasks.}
\label{fig:tian_scene}
\end{figure}

\begin{figure}[!ht]
\centering
\includegraphics[scale=0.48]{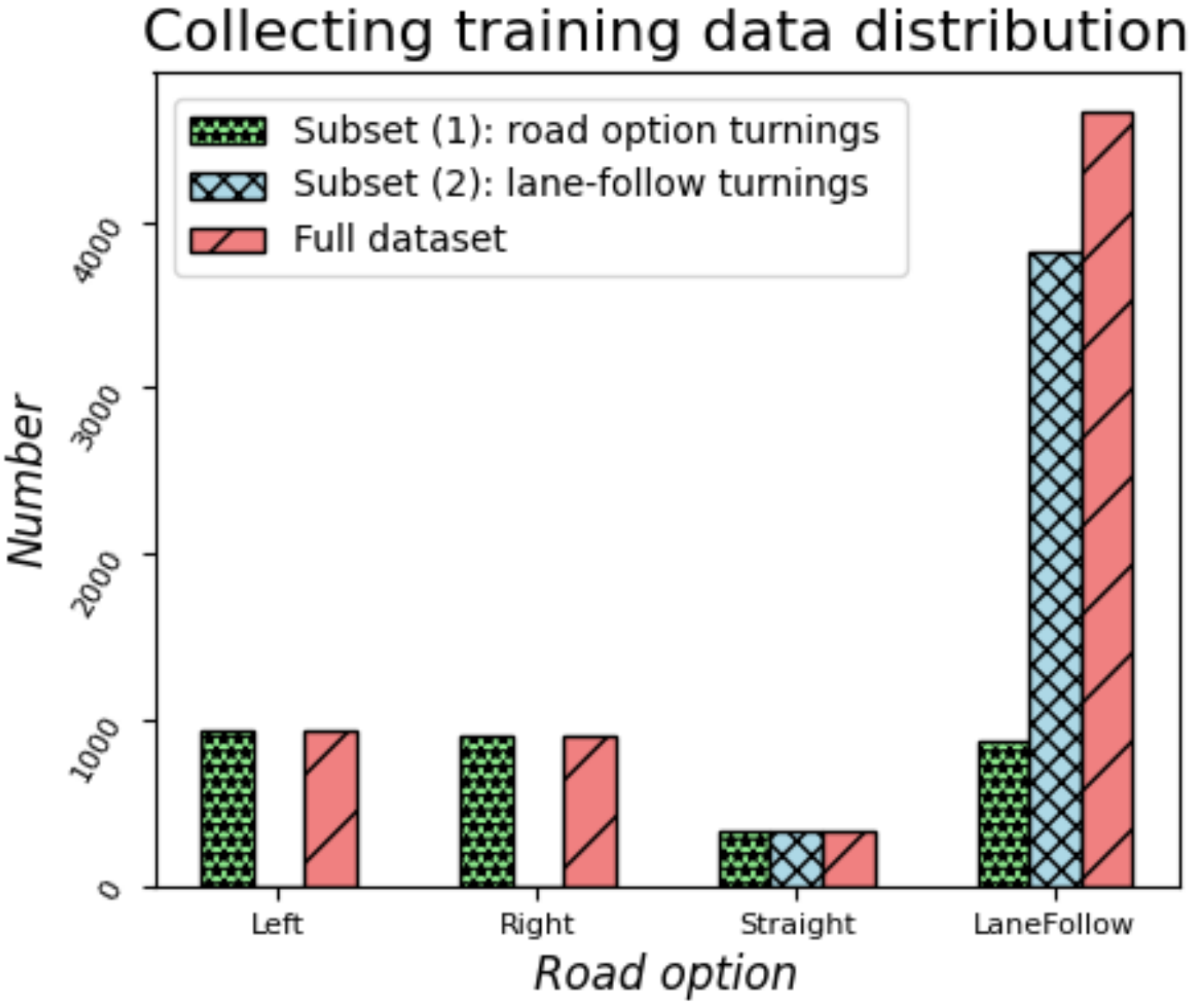}
\caption{Bar charts show the distribution of three training datasets. Subset (1) only contains the turning data of road option; Subset (2) only contains the turning data of lane keeping; The Full dataset is the union of two subsets.}
\label{fig:tian_dd}
\end{figure}

\subsubsection{Experimental Setup}
We compare the road option methods (XY-guided Imitation Learning, Yaw-guided Imitation Learning, Conditional Imitation Learning) in the ``square'' scene including the left and right turns of  the intersection, T-junction and lane-keeping, as shown in Figure \ref{fig:tian_scene}. There are many factors that affect the strategy, such as the initial value of the model parameters, the sampling order of the mini-batch. It is beyond the scope of this paper, there are detailed experimental analysis in \cite{Codevilla_2019_ICCV}. We consider the influence of data distribution on the strategy. We collect three different training datasets in the ``square'' scene of the Carla simulator including two extreme situations. Data distribution is shown in the bar of Figure \ref{fig:tian_dd}. 

\begin{itemize}
    \item Subset (1): only the turning data of the road option. 
    \item Subset (2): only the turning data of the lane keeping.
    \item Full dataset: union of the two subsets.
\end{itemize}

We use three datasets to train different road option models. Then, 32 turning tasks are tested by the models, as shown by the gray line in Figure \ref{fig:tian_scene} (a). And we count the success rate. Further, to verify the robustness of the methods, we add Gaussian noise to the models that $100\%$ complete the tasks  and test again.

\subsubsection{Experimental Results and Analysis}

\begin{table}[htbp] 
    \centering
	\caption{Number of successes}
    \begin{threeparttable}
    
    	\begin{tabular}{lcccc}
		\hline 
		Dataset & XY-guided & Yaw-guided & CIL & Carla-yaw IL\tnote{*}  \\
		\hline 
		Subset (1) &22&27&18&29 \\
		Subset (2) &9&12&8&11 \\
        Full dataset &26&32&32&32 \\
	    \hline
	    \end{tabular}
        \begin{tablenotes}
        
        \footnotesize
        \item[*]  To compare with yaw calculated by coordinate, yaw obtains directly from the Carla API.   
        \end{tablenotes}
    \end{threeparttable}
\end{table}

The success rate is shown in Table \uppercase\expandafter{\romannumeral1}. To compare with the Yaw-guided IL obtaining yaw by coordinate calculation, the Carla-yaw IL is added to the experiment, which obtains the yaw directly by Carla API. On the subset, the guided method has more successful times than the CIL. Especially, the models trained by Subset (1) are far better than those trained by Subset (2). The Yaw-guided IL has more successful times than the XY-guided IL. Therefore, we consider that: (1) the guided method has generalization ability when facing the situation of the road option, but the CIL does not. (2) Compared with the turning task of lane keeping, the turning task of road option is more difficult. (3) Compared with coordinates, the yaw guidance has a higher success rate. To further show point of view, we draw the trajectory of specific tasks in the Carla simulator. In Figure \ref{fig:tian_scene} (c) ``Turn left in lane-follow'', We use Subset (1) to train the model and test the left-turning task of lane following. The guided methods complete the task. On the contrary, CIL lacks the training data of lane following, and the corresponding branch network parameters cannot be optimized, resulting in failure. In the ``Turn right in the T-junction'', we use Subset (2) to train the model and test the right-turning of road option. Only Yaw-guided IL completes the task. XY-guided IL and CIL cannot generate the right-turning strategy by learning in the lane-following data, which leads to failure. Therefore, the Yaw guided IL is the best road option method and the turning of the road option is more difficult than the lane-keeping turning.

On the full dataset, both Yaw-guided IL and CIL complete all tasks. To compare the quality of task completion, we present a specific trajectory, in Figure \ref{fig:tian_scene} (d) ``Turn left in the intersection''. We use the full dataset to train models and test the left-turning task of road option. Yaw-guided IL and CIL complete the task, and the trajectory of CIL is closer to the standard trajectory. Therefore, compared with CIL, the guided method is more sensitive to the data distribution of left and right turns. XY-guided IL not only fails the task but also moves in the opposite direction. Because the guided data, as the input of the network, determines the behavior of the vehicle together with the image data. Obviously, the image data has a greater impact on vehicle behavior at this moment. 

\begin{table}[htbp]
    \centering
	\caption{Number of successes (Add Noise)}
	{
	\begin{tabular}{lccc}
		\toprule[1.2pt]
		Gaussian noise & N(0, $10^2$) & N(0, $20^2$) & N(0, $30^2$) \\
		\toprule[0.6pt]
		Yaw-guidance & $31.6\pm0.49$ & $30.4\pm0.80$ & $20.0\pm1.26$  \\
		Carla-yaw    & $31.2\pm0.40$ & $30.8\pm0.75$ & $18.4\pm1.02$  \\
		\toprule[1.2pt]
		Probability &3\%&5\%&10\% \\
		\toprule[0.6pt]
		CIL    & $24.6\pm7.58$ & $25.0\pm6.63$ & $20.4\pm10.29$  \\
	    \toprule[1.2pt]
	\end{tabular}
}
	\label{tab:noise}
\end{table}

In the real world, GPS navigation signals are disturbed by noise (for the guided IL), and users maybe issue wrong instructions (for the CIL). Therefore, the robustness of the road option method is very important for applying the model. We add noise to the models, Carla-yaw IL, yaw-guided IL and CIL, which complete all tasks. The guided method inputs the trajectory data into the network. Therefore, we directly add Gaussian noise to the raw guided data. Unlike the guided method, CIL selects different branch conditional networks by command. Therefore, we add five consecutive frames of wrong commands as noise based on the different probabilities. We choose the five different random numbers to repeat the experiments and calculate the mean and standard deviation of success times. The experimental results are shown in Table \ref{tab:noise}. Compared with the guided method, CIL is significantly affected by noise. Especially, when the wrong command is added with $10\%$ probability, the variance of the experimental results reaches $10.29^2$. With the increase of noise, the successful number of the guided methods gradually decreases, but the variance does not change much. Therefore, the guided method is more robust than the CIL.

\subsection{YILRatt in Urban Environments}

We set up experiments in Carla urban environments (dynamic obstacles). We train the YILRatt and CIL, CILRS models in the Town01 and test in the new scene Town02. Benchmark testing demonstrates the superior performance of the YILRatt architecture.

\subsubsection{Experimental Setup}

\begin{figure}[htbp]
\centering
\includegraphics[scale=0.4]{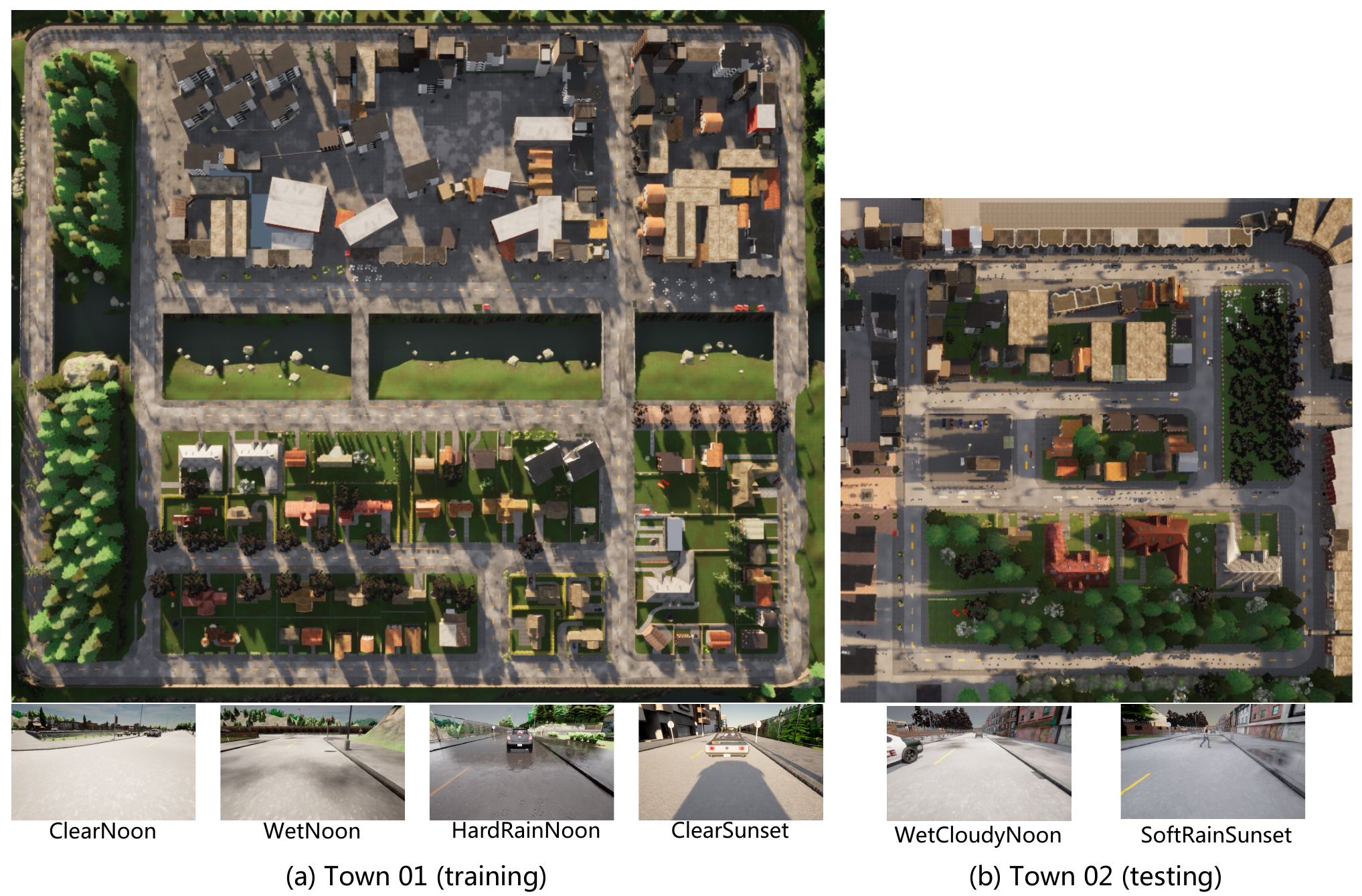}
\caption{The Carla urban environments. The data is collected in Town01 for training. Town02 is for testing. Views from onboard camera are based on the different weather conditions.}
\label{fig:t1_t2}
\end{figure}

\begin{figure}[htbp]
\centering
\includegraphics[scale=0.48]{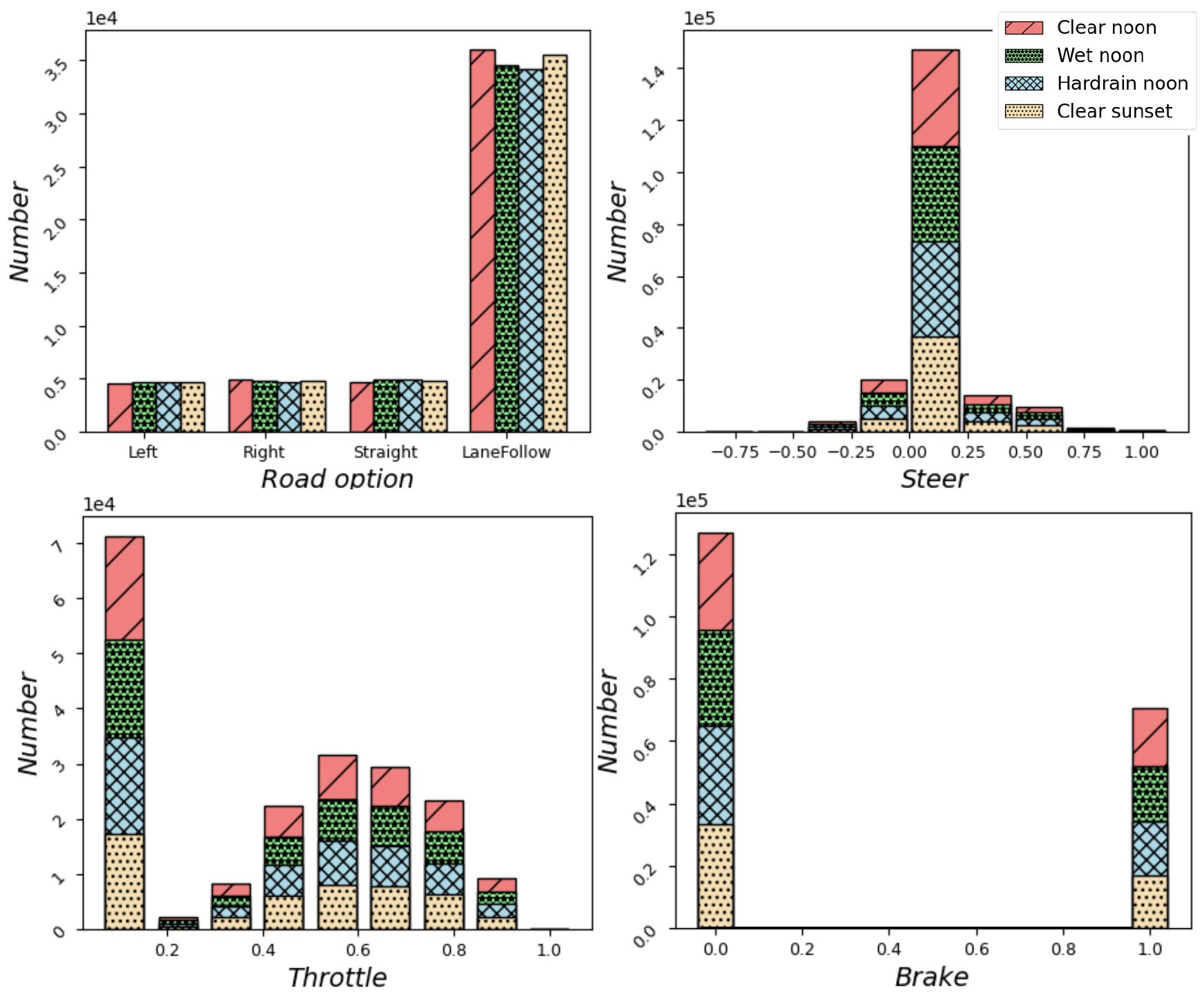}
\caption{The figure shows the distribution of training data under four weather conditions. The data is uniformly distributed under the road option (left, right and straight). The lane-follow data is about 7 times of others. Because the lane-follow data includes not only straight but also turns. The data distribution of steer, throttle, and brake are displayed in turn.}
\label{fig:data_dist}
\end{figure}

Dataset is the basis to ensure that the model completes the navigation task. We collect about \textbf{200} thousand data under four weather conditions (ClearNoon,  WetNoon, HardRainNoon, and ClearSunset) in Carla Town01, as shown in Figure \ref{fig:t1_t2} (a). The unmanned vehicle is controlled by the Carla AI algorithm to collect data. Control noise is added to $10\%$ of the data to improve the robustness of the model. But the recorded data is still the control signal. The dataset is divided into two parts: RGB images and labels. The size of the original image collected by the onboard camera is $3\times600\times800$ (channel, height, width). After cropping and scaling, the image size change to $3\times88\times200$. This size reduces the amount of calculation while ensuring that all environmental information is included. Labels include control (steer, throttle, and brake), road option (straight, left, right, and lane-follow), yaw and speed. The dataset is evenly distributed under the four weather conditions. However, lane keeping has far more data than the others. Because it has a high probability of appearing in the simulation scene, and contains turning data. The steer obeys the normal distribution. Throttle and brake are mutually exclusive, as shown in Figure \ref{fig:data_dist}. 

So far, \textit{CoRL2017} and \textit{NoCrash} are two commonly used benchmarks based on Carla 0.8.4. Refer to the original design standards, we have made improvements in the latest Carla 0.9.11. Benchmark is divided into two types of scenarios: \textit{Training condition} and \textit{New Town02 \& Weather} (WetCloudyNoon, SoftRainSunset). Navigation tasks are divided into five categories according to the degree of difficulty: Straight, One Turn, Navigation Empty, Nav. Regular (Vehicles: 30, Pedestrians: 10), Nav. Dense (Vehicles: 50, Pedestrians: 30). Each type of navigation task contains 25 paths, of which the route distance in Town01 is not less than 1.0 km, and the distance in Town02 is not less than 0.3 km. We use three metrics to measure the strategy: success rate, lane violations, and traffic light violations. We consider that the task is successful when the vehicle arrives at the destination without collision within the specified time. Especially, the violations of failed tasks are not included in the statistics.

All models are trained using the above dataset. The models use Adam optimizer with mini-batch 512 samples. The initial learning rate is 0.001. And every 10 epochs, the learning rate is reduced by half. The training process includes 100 epochs. The RGB image, speed and yaw are normalized processes and as inputs to the model. Image augmentation (blur, noise, brightness) are used to improve the generalization ability of the model. The outputs of the model are control and speed. The loss function is composed of the two parts and its weights are $\alpha=1.0$ and $\beta=0.1$ respectively. We record the model parameters that minimize the loss of the evaluation in 100 epochs and use them for testing the benchmark.

\subsubsection{Benchmark Testing}

\begin{table*}[htbp]
	\centering
	\caption{Navigation Success Rate}
	\resizebox{\textwidth}{!}
	{
	\begin{tabular}{lcccccccc}
		\toprule 
		& \multicolumn{4}{c}{Training Conditions} & \multicolumn{4}{c}{New Town02 \& Weather} \\
		Task & CIL & CILRS & YILRatt & YILRatt (pretrained) & CIL & CILRS & YILRatt & YILRatt(pretrained) \\
		\midrule 
		Straight &97&98&\textbf{99}&97&\textbf{100}&\textbf{100}&\textbf{100}&\textbf{100} \\
		One Turn &95&\textbf{100}&99&\textbf{100}&\textbf{100}&94&\textbf{100}&\textbf{100} \\
		Navigation (Empty)&41&50&\textbf{92}&86&64&44&88&\textbf{94} \\
		\midrule 
		Nav. (Regular) &$67.33\pm4.92$&$48.0\pm16.27$&$\bm{92.33\pm2.05}$&$92.0\pm2.16$&$46.67\pm1.89$&$35.33\pm6.18$&$76.66\pm0.94$&$\bm{87.33\pm6.80}$ \\
		Nav. (Dense) &$55.33\pm2.49$&$66.67\pm0.94$&$88.67\pm1.70$&$\bm{93.67\pm2.62}$&$25.33\pm5.00$&$34.0\pm4.32$&$76.0\pm3.27$&$\bm{82.67\pm0.94}$ \\
	    \bottomrule
	\end{tabular}
	}
	\label{tab:benchmark}
\end{table*}

 \begin{table*}[htbp]
	\centering
	\caption{Lane and Traffic light violations}
	\resizebox{0.8\textwidth}{!}
	{
	\begin{tabular}{lccccc}
		\toprule 
		\multicolumn{2}{l}{\multirow{2}*{Task}}&\multicolumn{4}{c}{New Town02 \& Weather}  \\
		\multicolumn{2}{c}{~} & CIL & CILRS & YILRatt & YILRatt (pretrained) \\
		\midrule 
		\multirow{2}*{\makecell[l]{Lane Violation \\(Infraction per km)}} & Nav. (Regular) &$4.70\pm0.22$&$4.65\pm0.30$&$3.23\pm0.36$&\bm{$2.91\pm0.93$}\\
		~ & Nav. (Dense) &$4.53\pm0.54$&$4.42\pm0.39$&$3.38\pm0.04$&\bm{$1.94\pm0.01$} \\
		\midrule 
		\multirow{2}*{\makecell[l]{Traffic Light Violation \\(Violations/the total number)} } & Nav. (Regular) &$0.50\pm0.03$&$0.52\pm0.08$&\bm{$0.31\pm0.04$}&$0.36\pm0.04$\\
		~ & Nav. (Dense) &$0.63\pm0.27$&$0.35\pm0.14$&\bm{$0.34\pm0.07$}&$0.35\pm0.10$ \\
	    \bottomrule
	\end{tabular}
}
	\label{tab:invade}
\end{table*}

\begin{figure*}[htbp]
\centering
\includegraphics[scale=0.39]{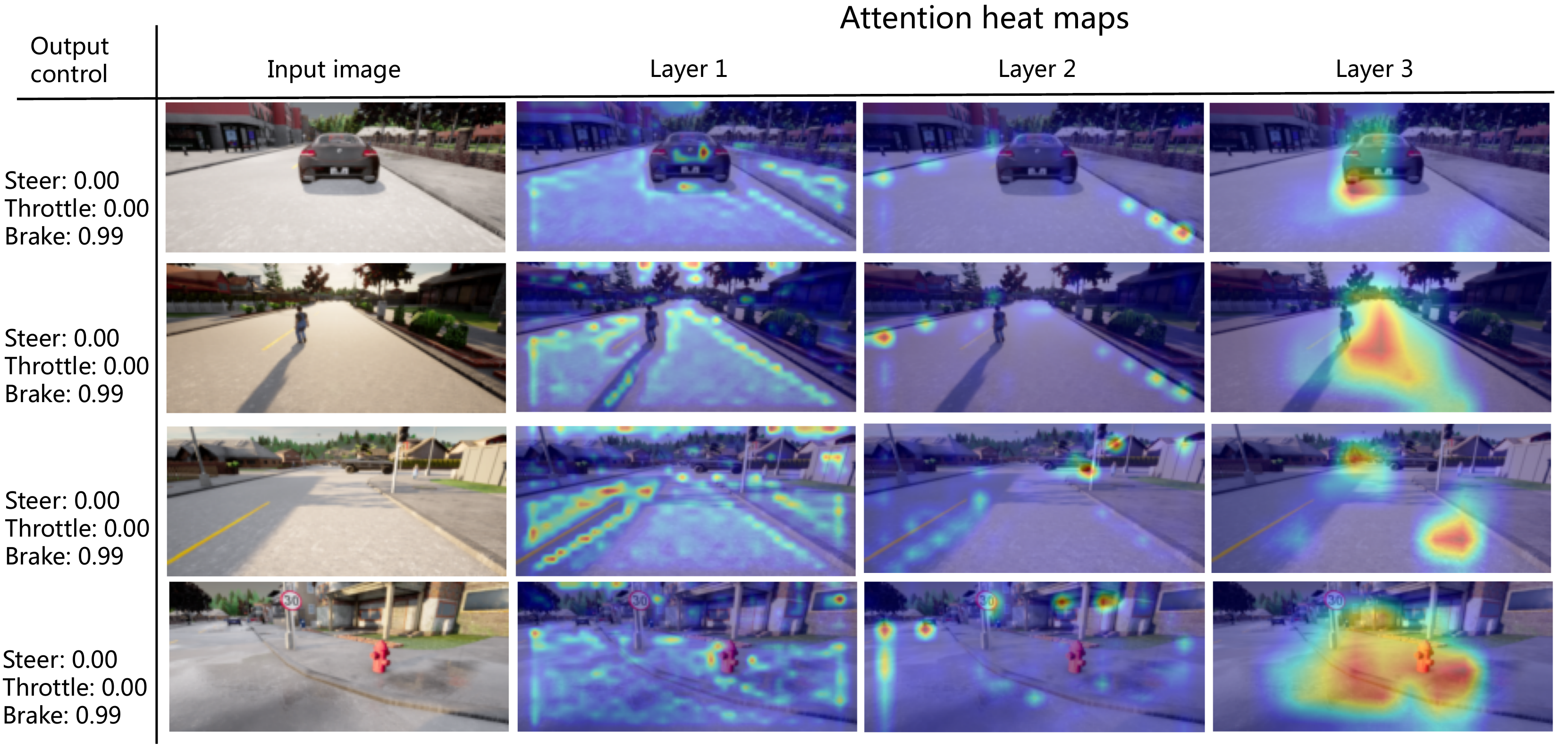}
\caption{The attention map is derived from the YILRatt(pretrained). Four sets of images focus sharply on the objects obstructing the movement of vehicles.}
\label{fig:att_map}
\end{figure*}

\begin{figure*}[htbp]
\centering
\includegraphics[scale=0.59]{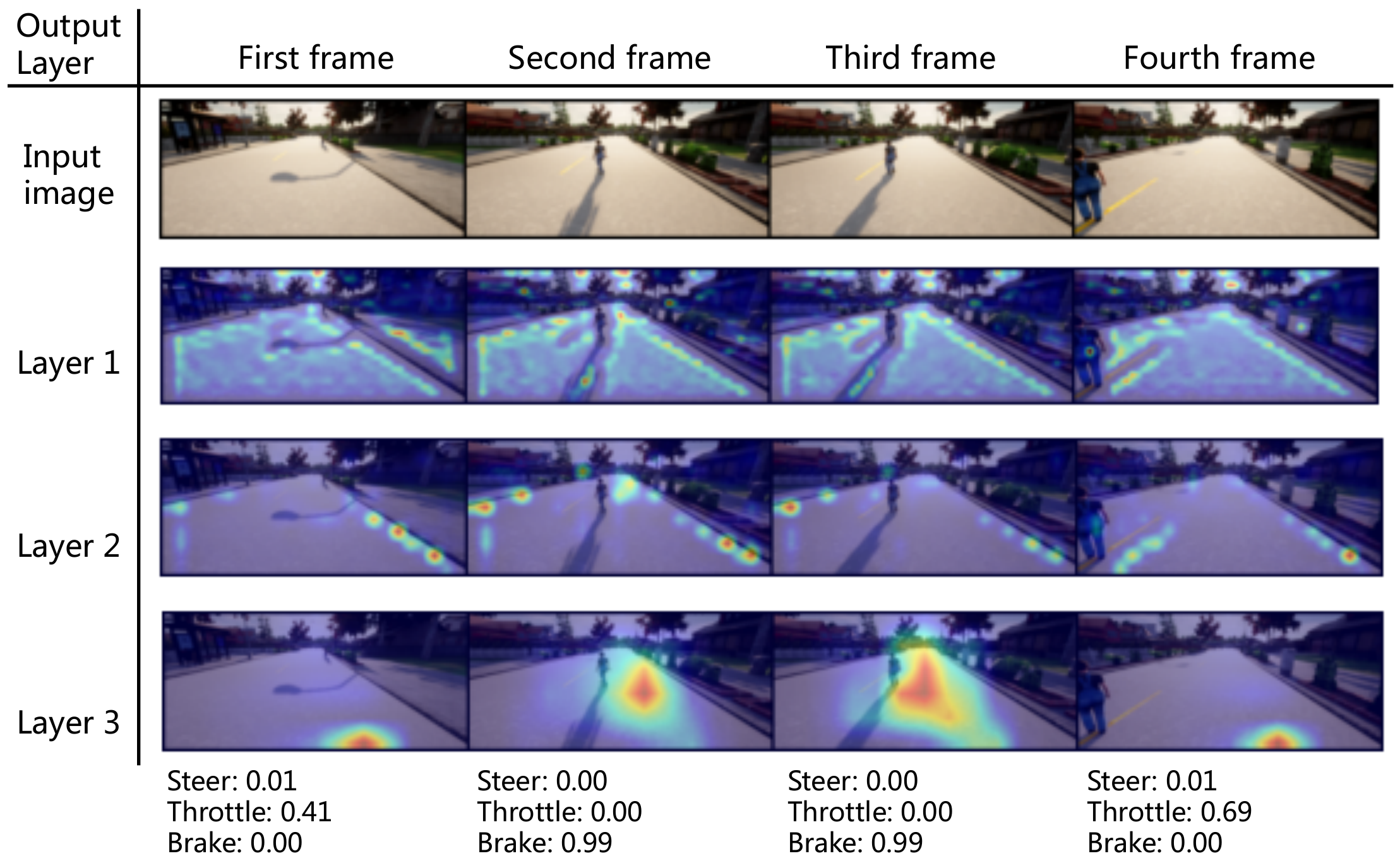}
\caption{Heat maps are four consecutive frames. It demonstrates the process of the vehicle avoiding a pedestrian, slowing down, and then accelerating.}
\label{fig:att_map_pede}
\end{figure*}
We test four methods, the CIL, CILRS, YILRatt and the pretrained model by Pytorch ResNet34, on the benchmark. These end-to-end imitation learning methods only use a fixed dataset for training. No additional auxiliary information is required. We show the experimental results in Table \ref{tab:benchmark}. The result shows the navigation success rates of the four algorithms in static and dynamic scenarios. Static scenes include \textit{Straight}, \textit{One Turn}, \textit{Navigation (empty)}. Each algorithm selects the best seed results of five runs. Dynamic scenes include \textit{Regular} and \textit{Dense}. Mean and standard are derived from the three runs. In static scenarios, all methods have a high success rate in \textbf{Training Conditions} and generalization ability in \textbf{New Town02 \& Weather}. But when there are dynamic obstacles in the scene, the success rate of CIL and CILRS drops significantly. In particular, under \textbf{New Town02 \& Weather}, CIL has a higher success rate than CILRS. It shows that when the training data is insufficient, the large-scale network will appear over-fitting. The attention mechanism obviously improves the generalization of the model. In the regular and dense conditions, the success rate of YILRatt (pretrained) is increased by $52\%$ and $48.67\%$ respectively than that of CILRS. The experimental results also show that the pretrained model is helpful to strategy optimization, but it is not obvious.

We count the number of lane and red traffic light violations in Table \ref{tab:invade}. We do statistics under the \textbf{New Town02 \& Weather} dynamic conditions. Mean and standard are derived from the three runs. YILRatt (pretrained) has the lowest number of lane violations ($2.91\pm0.93$ in \textit{Regular} and $1.94\pm0.01$ in \textit{Dense}). And in terms of signal light recognition, YILRatt performs best, ($0.31\pm0.04$ in \textit{Regular} and $0.34\pm0.07$ in \textit{Dense}). The possible reason why YILRatt is better than pretrained model is: \textbf{ImageNet} dataset is used to pretrain ResNet34, which has no obvious effect on traffic lights recognition. Experimental data shows that the end-to-end imitation learning method we proposed can perform best not only to complete the task quantity but also in quality. Therefore, the YILRatt can well integrate the perception and control modules to realize end-to-end autonomous driving.

\subsubsection{Attention Heat Maps}

On the one hand, the Attention module improves the perception ability of the model by fusing local and global features. On the other hand, we use the attention heat maps to understand the causal relationship between the scene and decision. Especially, maps help us analyze the reason of failure cases. We superimpose the feature map after attention transform with the raw image with weight value to get the heat map. We select the pedestrian, the vehicle, the traffic light, and fire hydrant to demonstrate in Figure \ref{fig:att_map}. The heat maps of Layer 1 pay attention to the passable area of the vehicle. The heat maps of Layer 2 pay attention to lane line characteristics. The heat maps of Layer 3 pay attention to details. The vehicle behavior corresponding to the scene is braking. However, in the hydrant case, there was an unpredictable result. The vehicle mistakes the fire hydrant for an obstacle to avoid and makes a decision to stop, which leads to the failure of the task.

Furthermore, we analyze the scene of the vehicle avoiding a pedestrian through sequential frame heat maps in Figure \ref{fig:att_map_pede}. When the pedestrian is far away from the vehicle, the vehicle pays attention to the nearby passable area and maintains the speed (throttle=0.41). When a pedestrian obstructs the vehicle, the vehicle pays attention to the pedestrian and brakes (brake=0.99). After the pedestrian passes, the vehicle again shifts its attention to the passable area and accelerates (throttle=0.69). By attention heat maps, we visualize environmental characteristics and clarify the driving behavior of the unmanned vehicle. For more case studies, please see the supplementary video  \href{https://youtu.be/EQg3ZPHTi48}{https://youtu.be/EQg3ZPHTi48} .

\section{Conclusion}\label{section5:conclusion}
The experimental data collection, benchmark testing, and result analysis show that our designed end-to-end imitation learning architecture YILRatt was better than the SOTA CILRS. The yaw-guided method achieves road option and improves data utilization. The perception module based on ResNet34, fusing attention mechanism improves the accuracy of scene recognition. Attention heat maps explain the reason for making decisions. 

Nonetheless, from the analysis of failure cases, it can be seen that the YILRatt cannot handle sequence information. Models with timing processing capabilities, such as Transformer, can be considered. What follows is the increase of model parameters, we need to collect more data to train the model to prevent overfitting. Another problem is that the IL has an expert strategy bottleneck. Reinforcement learning is a good attempt to explore better strategies.

\bibliographystyle{IEEEtran}
\bibliography{YILRatt_AD_intial_20211102}

\end{document}